\def\year{2018}\relax
\documentclass[letterpaper]{article} 
\usepackage{aaai18}  
\usepackage{times}  
\usepackage{helvet}  
\usepackage{courier}  
\usepackage{url}  
\usepackage{graphicx}  

\usepackage{amsfonts}       
\usepackage{nicefrac}       
\usepackage[labelfont=bf]{caption}
\usepackage{etoolbox,amssymb,amsmath}
\usepackage{hyperref}
\usepackage{enumerate}
\usepackage{mdframed}
\usepackage{stmaryrd}
\usepackage[mathscr]{euscript}
\graphicspath{ {images/} }

\newcommand{\tensor}[1]{\boldsymbol{\mathscr{#1}}}

\newcommand{\U}[0]{\mathbf{U}}

\newcommand{\createbigfig}[3]{
\begin{figure*}[ht]
  \begin{mdframed}
    \centering
    \includegraphics[width=#3\linewidth]{{#1}.png} 
    \caption{{#2}}
    \label{fig:#1}
  \end{mdframed}
\end{figure*}
}
\DeclareMathOperator*{\argmin}{argmin}
\nocopyright
 
\begin{document}
%
\title{Word Embeddings via Tensor Factorization}

\author{ Eric Bailey, Charles Meyer, and Shuchin Aeron \\ Tufts University, Medford, MA 02215} 

\maketitle
\begin{abstract}
Many state-of-the-art word embedding techniques involve factorization of a co-occurrence based matrix.
We aim to extend this approach by studying word embedding techniques that involve factorization of co-occurrence based \textit{tensors} ($N$-way arrays).
We present two new word embedding techniques based on tensor factorization and show that they outperform common methods on several semantic NLP tasks when given the same data.
To train one of the embeddings, we present a new joint tensor factorization problem and an approach for solving it.
Furthermore, we modify the performance metrics for the Outlier Detection \cite{outlier1} task to measure the quality of higher-order relationships that a word embedding captures. 
Our tensor-based methods significantly outperform existing methods at this task when using our new metric.
Finally, we demonstrate that vectors in our embeddings can be composed \textit{multiplicatively} to create different vector representations for each meaning of a polysemous word. We show that this property stems from the higher order information that the vectors contain, and thus is unique to our tensor based embeddings.

\end{abstract}

    \section{Introduction} \label{intro}
   Word embeddings have been used to improve the performance of many NLP tasks including language modelling \cite{Bengio:2003:NPL:944919.944966}, machine translation \cite{DBLP:journals/corr/BahdanauCB14}, and sentiment analysis \cite{yoonkim}.
The broad applicability of word embeddings to NLP implies that improvements to their quality will likely have widespread benefits for the field.

The word embedding problem is to learn a mapping $\eta: V\rightarrow \mathbb{R}^k$ ($k \approx$ 100-300 in most applications) that encodes meaningful semantic and/or syntactic information. 
For instance, in many word embeddings, $\eta($car$) \approx \eta($truck$)$, since the words are semantically similar. 

More complex relationships than similarity can also be encoded in word embeddings. For example, we can answer analogy queries of the form $a : b :: c :$ ? using simple arithmetic in many state-of-the-art embeddings \cite{word2vec}. The answer to bed $:$ sleep $::$ chair $:$ $x$ is given by the word whose vector representation is closest to $\eta($sleep$) - \eta($bed$) + \eta($chair$)$ ($\approx \eta($sit$)$). Other embeddings may encode such information in a nonlinear way \cite{simplesupervised}.

\cite{word2vec} demonstrates the additive compositionality of their \texttt{word2vec} vectors: one can sum vectors produced by their embedding to compute vectors for certain \textit{phrases} rather than just vectors for words. Later in this paper, we will show that our embeddings naturally give rise to a form of \textit{multiplicative} compositionality that has not yet been explored in the literature.

Almost all recent word embeddings rely on the distributional hypothesis \cite{harris54}, which states that a word's meaning can be inferred from the words that tend to surround it.
To utilize the distributional hypothesis, many embeddings are given by a low-rank factor of a matrix derived from co-occurrences in a large unsupervised corpus, see \cite{glove,nnse,matrixfact} and \cite{matrix_negsample}. 

Approaches that rely on matrix factorization only utilize pairwise co-occurrence information in the corpus. We aim to extend this approach by creating word embeddings given by factors of \textit{tensors} containing higher order co-occurrence data. 


\subsection{Related work} \label{relatedwork}
Some common word embeddings related to co-occurrence based matrix factorization include GloVe \cite{glove}, \texttt{word2vec} \cite{matrixfact}, LexVec \cite{matrix_negsample}, and NNSE \cite{nnse}.
In contrast, our work studies word embeddings given by factorization of tensors.
An overview of tensor factorization methods is given in \cite{tensorfact}. 

Our work uses factorization of symmetric nonnegative tensors, which has been studied in the past \cite{successive,symmetrictensorrank}.
In general, factorization of tensors has been applied to NLP in \cite{van2013tensor} and factorization of nonnegative tensors  \cite{VandeCruys:2009:NTF:1705415.1705426}. 
Recently, factorization of symmetric tensors has been used to create a generic word embedding \cite{orthcp} but the idea was not explored extensively. 
Our work studies this idea in much greater detail, fully demonstrating the viability of tensor factorization as a technique for training word embeddings. 

Composition of word vectors to create novel representations has been studied in depth, including additive, multiplicative, and tensor-based methods \cite{compositionality_overview,blacoe_lapata_composition2}.
Typically, composition is used to create vectors that represent \textit{phrases} or \textit{sentences}.
Our work, instead, shows that pairs of word vectors can be composed multiplicatively to create different vector representations for the various meanings of a \textit{single} polysemous word.
    
    \section{Mathematical preliminaries} \label{prelims}
  \subsection{Notation}
Throughout this paper we will write 
scalars in lowercase italics $\alpha$,
vectors in lowercase bold letters $\mathbf{v}$,
matrices with uppercase bold letters $\mathbf{M}$, and
tensors (of order $N>2$) with Euler script notation $\tensor{X}$, as is standard in the literature. 

\subsection{Pointwise Mutual Information}  \label{ppmi}

Pointwise mutual information (PMI) is a useful property in NLP that quantifies the likelihood that two words co-occur \cite{matrixfact}. It is defined as: \begin{align*}
    PMI(x,y) &= \text{log}\cfrac{p(x,y)}{p(x)p(y)}
\end{align*}

where $p(x,y)$ is the probability that $x$ and $y$ occur together in a given fixed-length context window in the corpus, irrespective of order.
 
It is often useful to consider the \textit{positive} PMI (PPMI), defined as:
\[
PPMI(x,y) := \text{max}(0, PMI(x,y))
\]
since negative PMI values have little grounded interpretation \cite{ppmi_stats,matrixfact,VandeCruys:2009:NTF:1705415.1705426}.

Given an indexed vocabulary $V = \{w_1, \dots, w_{|V|}\}$, one can construct a $|V| \times |V|$ PPMI matrix $\mathbf{M}$ where $m_{ij} = PPMI(w_i, w_j)$. Many existing word embedding techniques involve factorizing this PPMI matrix \cite{matrixfact,nnse,matrix_negsample}.

PMI can be generalized to $N$ variables. While there are many ways to do so \cite{ppmi}, in this paper we use the form defined by:
\begin{align*}
PMI(x_1^N) &= \text{log}\cfrac{p(x_1, \dots, x_N)}{p(x_1) \cdots p(x_N)}
\end{align*}
where $p(x_1, \dots, x_N)$ is the probability that \textit{all} of $x_1, \dots, x_N$ occur together in a given fixed-length context window in the corpus, irrespective of their order.

In this paper we study 3-way PPMI \textit{tensors} $\tensor{M}$, where $m_{ijk} = PPMI(w_i, w_j, w_k)$, as this is the natural higher-order generalization of the PPMI matrix.
We leave the study of creating word embeddings with $N$-dimensional PPMI tensors ($N>3$) to future work.

\subsection{Tensor factorization} \label{tensorfact}
Just as the rank-$R$ matrix decomposition is defined to be the product of two factor matrices ($\mathbf{M} \approx \U \mathbf{V}^\top$), the canonical rank-$R$ tensor decomposition for a third order tensor is defined to be the product of three factor matrices \cite{tensorfact}:
\begin{align}
\tensor{X} &\approx \sum_{r=1}^{R} \mathbf{u}_r \otimes \mathbf{v}_r \otimes \mathbf{w}_r =: \llbracket \mathbf{U}, \mathbf{V}, \mathbf{W}   \rrbracket,
\end{align}
where $\otimes$ is the outer product: $(\mathbf{a} \otimes \mathbf{b} \otimes \mathbf{c})_{ijk} = a_i b_j c_k$.
This is also commonly referred to as the rank-\textit{R CP Decomposition}.
Elementwise, this is written as:
\[
x_{ijk} \approx \sum_{r=1}^R u_{ir} v_{jr} w_{kr} = \langle \mathbf{u}_{:,i} * \mathbf{v}_{:,j}, \mathbf{w}_{:,k} \rangle,
\]
where $*$ is elementwise vector multiplication and $\mathbf{u}_{:,i}$ is the $i^{th}$ row of $\U$.
In our later section on multiplicative compositionality, we will see this formulation gives rise to a meaningful interpretation of the elementwise product between vectors in our word embeddings.


\textbf{Symmetric CP Decomposition.}
In this paper, we will consider \textit{symmetric} CP decomposition of \textit{nonnegative} tensors \cite{symnonneg_exists,tensorfact}. Since our $N$-way PPMI is nonnegative and invariant under permutation, the PPMI tensor $\tensor{M}$ is nonnegative and supersymmetric, i.e. $m_{ijk} = m_{\sigma(i)\sigma(j)\sigma(k)} \geq 0$ for any permutation $\sigma \in S_3$.

In the symmetric CP decomposition, instead of factorizing $\tensor{M} \approx \llbracket \U, \mathbf{V}, \mathbf{W} \rrbracket$, we factorize $\tensor{M}$ as the triple product of a \textit{single} factor matrix $\mathbf{U} \in \mathbb{R}^{|V| \times R}$ such that 
\[
\tensor{M} \approx \llbracket \mathbf{U}, \mathbf{U}, \mathbf{U} \rrbracket
\]

In this formulation, we use $\U$ to be the word embedding so the vector for $w_i$ is the $i^{th}$ row of $\U$  similar to the formulations in \cite{matrixfact,nnse,glove}.

It is known that the optimal rank-$k$ CP decomposition exists for symmetric nonnegative tensors such as the PPMI tensor \cite{symnonneg_exists}.
However, finding such a decomposition is NP hard in general \cite{cp_np} so we must consider approximate methods.



In this work, we only consider the symmetric CP decomposition, leaving the study of other tensor decompositions (such as the Tensor Train or HOSVD \cite{tensortrain,tensorfact}) to future work.
    
    \section{Methodologies}  \label{methodology}
\subsection{Computing the Symmetric CP Decomposition}
The $\Theta(|V|^3)$ size of the third order PPMI tensor presents a number of computational challenges. In practice, $|V|$ can vary from $10^4$ to $10^6$, resulting in a tensor whose naive representation requires at least $4*10,000^3$ bytes = $4$ TB of floats. 
Even the sparse representation of the tensor takes up such a large fraction of memory that standard algorithms such as successive rank-1 approximation \cite{successive,successive_nodeco} and alternating least-squares \cite{tensorfact} are infeasible for our uses.
Thus, in this paper we will consider a stochastic online formulation similar to that of \cite{online_cp_decomp}.

We optimize the CP decomposition in an online fashion, using small random subsets $\tensor{M}^t$ of the nonzero tensor entries to update the decomposition at time $t$.
In this minibatch setting, we optimize the decomposition based on the current minibatch and the previous decomposition at time $t-1$.
To update $\U$ (and thus the symmetric decomposition), we first define a decomposition loss $\boldsymbol{\mathscr{L}}(\tensor{M}^t, \U)$ and minimize this loss with respect to $\U$ using Adam \cite{adam}.

At each time $t$, we take $\tensor{M}^t$ to be all co-occurrence triples (weighted by PPMI) in a fixed number of sentences (around 1,000) from the corpus.
We continue training until we have depleted the entire corpus.

For $\tensor{M}^t$ to accurately model $\tensor{M}$, we also include a certain proportion of elements with zero PPMI (or ``negative samples'') in $\tensor{M}^t$, similar to that of \cite{matrix_negsample}. 
We use an empirically found proportion of negative samples for training, and leave discovery of the optimal negative sample proportion to future work.

\subsection{Word Embedding Proposals}
\textbf{CP-S.}
The first embedding we propose is based on symmetic CP decomposition of the PPMI tensor $\tensor{M}$ as discussed in the mathematical preliminaries section.
The optimal setting for the word embedding $\mathbf{W}$ is:
\[
\mathbf{W} := \argmin_\U  || \tensor{M} - \llbracket \U, \U, \U \rrbracket ||_F
\]
Since we cannot feasibly compute this exactly, we minimize the loss function defined as the squared error between the values in $\tensor{M}^t$ and their predicted values:
\begin{align*}
\boldsymbol{\mathscr{L}}(\tensor{M}^t, \U) 
&= \sum_{m^t_{ijk} \in \tensor{M}^t}
({m}^t_{ijk} - \sum_{r=1}^{R} u_{ir}u_{jr}u_{kr})^2
\end{align*}
using the techniques discussed in the previous section. 

\textbf{JCP-S.} \label{jcp-s}
A potential problem with CP-S is that it is \textit{only} trained on third order information.
To rectify this issue, we propose a novel joint tensor factorization problem we call \textit{Joint Symmetric Rank-$R$ CP Decomposition}.
In this problem, the input is the fixed rank $R$ and a list of supersymmetric tensors $\tensor{M}_{n}$ of different orders but whose axis lengths all equal $|V|$.
Each tensor $\tensor{M}_{n}$ is to be factorized via rank-$R$ symmetric CP decomposition using a \textit{single} $|V| \times R$ factor matrix $\U$.

To produce a solution, we first define the loss at time $t$ to be the sum of the reconstruction losses of each different tensor:
\[
\boldsymbol{\mathscr{L}}_{\text{joint}} ((\tensor{M}^t)_{n=2}^{N}, \U) =
\sum_{n=2}^{N} \boldsymbol{\mathscr{L}}(\tensor{M}_n^t, \U),
\]
where $\tensor{M}_n$ is an $n$-dimensional supersymmetric PPMI tensor. We then minimize the loss with respect to $\U$. 
Since we are using at most third order tensors in this work, we assign our word embedding $\mathbf{W}$ to be:
\begin{align*}
\mathbf{W} &:= 
\argmin_\U \boldsymbol{\mathscr{L}}_{\text{joint}} (\mathbf{M}_2, \tensor{M}_3, \U) \\
&= 
\argmin_\U \bigg[ \boldsymbol{\mathscr{L}}(\mathbf{M}_2, \U) +  \boldsymbol{\mathscr{L}}(\tensor{M}_3, \U) \bigg]
\end{align*}

This problem is a specific instance of Coupled Tensor Decomposition, which has been studied in the past \cite{jointtensor1,jointtensor2}. In this problem, the goal is to factorize multiple tensors using \textit{at least} one factor matrix in common.
A similar formulation to our problem can be found in \cite{polynomial_joint_sym_cp}, which studies blind source separation using the algebraic geometric aspects of jointly factorizing numerous supersymmetric tensors (to unknown rank). In contrast to our work, they outline some generic rank properties of such a decomposition rather than attacking the problem numerically. Also, in our formulation the rank is fixed and an approximate solution must be found.
Exploring the connection between the theoretical aspects of joint decomposition and quality of word embeddings would be an interesting avenue for future work.

To the best of our knowledge this is the first study of Joint \textit{Symmetric Rank-$R$} CP Decomposition.

\subsection{Shifted PMI}
In the same way \cite{matrixfact} considers factorization of positive shifted PMI matrices, we consider factorization of positive shifted PMI \textit{tensors} $\tensor{M}$, where $m_{ijk} = \max(PMI(w_i, w_j, w_k) - \alpha, 0)$ for some constant shift $\alpha$. We empirically found that different levels of shifting resulted in different qualities of word embeddings -- the best shift we found for CP-S was a shift of $\alpha \approx 3$, whereas any nonzero shift for JCP-S resulted in a worse embedding across the board. 
When we discuss evaluation we report the results given by factorization of the PPMI tensors shifted by the best value we found for each specific embedding.

\subsection{Computational notes}
When considering going from two dimensions to three, it is perhaps necessary to discuss the computational issues in such a problem size increase. However, it should be noted that the creation of pre-trained embeddings can be seen as a pre-processing step for many future NLP tasks, so if the training can be completed once, it can be used forever thereafter without having to take training time into account. 
Despite this, we found that the training of our embeddings was not considerably slower than the training of order-2 equivalents such as SGNS. 
Explicitly, our GPU trained CBOW vectors (using the experimental settings found below) in 3568 seconds, whereas training CP-S and JCP-S took 6786 and 8686 seconds respectively. 
    
    \section{Evaluation}  \label{evaluation}
In this section we present a quantitative evaluation comparing our embeddings to an informationless embedding and two strong baselines.
Our baselines are: 
\begin{enumerate}[1.]
    \item \textbf{Random} (random vectors with I.I.D. entries normally distributed with mean 0 and variance $\frac{1}{2}$), for comparing against a model with no meaningful information encoded
    \item \textbf{{word2vec}} (CBOW) \cite{word2vec}, for comparison against the most commonly used embedding method, as well as for comparison against a technique related to PPMI matrix factorization \cite{matrixfact}
    \item \textbf{NNSE}\footnote{The input to NNSE is an $m \times n$ matrix, where there are $m$ words and $n$ co-occurrence patterns. In our experiments, we set $m=n=|V|$ and set the co-occurrence information to be the number of times $w_i$ appears within a window of 5 words of $w_j$. As stated in the paper, the matrix entries are weighted by PPMI.} \cite{nnse}, for comparison against a technique that relies on an explicit PPMI matrix factorization
\end{enumerate}

For a fair comparison, we trained each model on the same corpus of 10 million sentences gathered from Wikipedia. We removed stopwords and words appearing fewer than 2,000 times (130 million tokens total) to reduce noise and uninformative words. Our word2vec and NNSE baselines were trained using the recommended hyperparameters from their original publications, and all optimizers were using using the default settings. Hyperparameters are always consistent across evaluations.

Because of the dataset size, the results shown should be considered a proof of concept rather than an objective comparison to state-of-the-art pre-trained embeddings. Due to the natural computational challenges arising from working with tensors, we leave creation of a full-scale production ready embedding based on tensor factorization to future work. 

As is common in the literature \cite{word2vec,nnse}, we use  300-dimensional vectors for our embeddings and all word vectors are normalized to unit length prior to evaluation.

\subsection{Quantitative tasks}
\textbf{Outlier Detection.}
The Outlier Detection task \cite{outlier1} is to determine which word in a list $L$ of $n + 1$ words is unrelated to the other $n$ which were chosen to be related. For each $w \in L$, one can compute its compactness score $c(w)$, which is the compactness of $L \setminus \{w\}$.
$c(w)$ is explicitly computed as the mean similarity of all word \textit{pairs} $(w_i, w_j) : w_i, w_j \in L \setminus \{w\}$.
The predicted outlier is $\text{argmax}_{w \in L} c(w)$, as the $n$ related words should form a compact cluster with high mean similarity. 

We use the WikiSem500 dataset \cite{outlier2} which includes sets of $n=8$ words per group gathered based on semantic similarity. Thus, performance on this task is correlated with the amount of semantic information encoded in a word embedding. Performance on this dataset was shown to be well-correlated with performance at the common NLP task of sentiment analysis \cite{outlier2}.

The two metrics associated with this task are accuracy and Outlier Position Percentage (OPP). Accuracy is the fraction of cases in which the true outlier correctly had the highest compactness score. OPP measures how \textit{close} the true outlier was to having the highest compactness score, rewarding embeddings more for predicting the outlier to be in $2^{nd}$ place rather than $n^{th}$ when sorting the words by their compactness score $c(w)$.

\textbf{3-way Outlier Detection.}
As our tensor-based embeddings encode higher order relationships between words, we introduce a new way to compute $c(w)$ based on \textit{groups of 3} words rather than pairs of words. We define the compactness score for a word $w$ to be:
\[
c(w) = \sum_{\mathbf{v}_{i_1} \neq \mathbf{v}_w} \sum_{\mathbf{v}_{i_2} \neq \mathbf{v}_w, \mathbf{v}_{i_1}} \sum_{\mathbf{v}_{i_3} \neq \mathbf{v}_w, \mathbf{v}_{i_1}, \mathbf{v}_{i_2}} sim(\mathbf{v}_{i_1}, \mathbf{v}_{i_2}, \mathbf{v}_{i_3}),
\]
where $sim(\cdot)$ denotes similarity between a group of 3 vectors. $sim(\cdot)$ is defined as:
\[
sim(\mathbf{v}_1, \mathbf{v}_2, \mathbf{v}_3) = \bigg(\frac{1}{3} \sum_{i=1}^3 ||\mathbf{v}_i - \frac{1}{3} \sum_{j=1}^3 \mathbf{v}_j||_2\bigg)^{-1}
\]
We call this evaluation method OD$3$.

The purpose of OD$3$ is to evaluate the extent to which an embedding captures $3^{rd}$ order relationships between words. As we will see in the results of our quantitative experiments, our tensor methods outperform the baselines on OD3, which validates our approach. 

This approach can easily be generalized to OD$N$ $(N > 3)$, but again we leave the study of higher order relationships to future work.

\textbf{Simple supervised tasks.} \label{simplesupervised_section}
\cite{simplesupervised} points out that the primary application of word embeddings is \textit{transfer learning} to NLP tasks. They argue that to evaluate an embedding's ability to transfer information to a relevant task, one must measure the embedding's \textit{accessibility of information} for actual downstream tasks. To do so, one must cite the performance of simple supervised tasks as training set size increases, which is commonly done in transfer learning evaluation \cite{simplesupervised}. 
If an algorithm using a word embedding performs well with just a small amount of training data, then the information encoded in the embedding is easily accessible.

The simple supervised downstream tasks we use to evaluate the embeddings are as follows:

\begin{enumerate}[1.]
    \item \textbf{Supervised Analogy Recovery.}
    We consider the task of solving queries of the form \textit{a : b :: c : }\textit{?} using a simple neural network as suggested in \cite{simplesupervised}.    
    The analogy dataset we use is from the Google analogy testbed \cite{word2vec}.
    
    \item \textbf{Sentiment analysis.}
    We also consider sentiment analysis as described by \cite{unsupervised_evaluation}.     
    We use the suggested Large Movie Review dataset \cite{large_sentiment_dataset}, containing 50,000 movie reviews.
    \end{enumerate}
All code is implemented using scikit-learn 
or TensorFlow 
and uses the suggested train/test split.

\textbf{Word similarity.}
To standardize our evaluation methodology, we evaluate the embeddings using word similarity on the common MEN and MTurk datasets \cite{MEN,MTurk}. For an overview of word similarity evaluation, see \cite{unsupervised_evaluation}.

\subsection{Quantitative results} \label{quantresults}

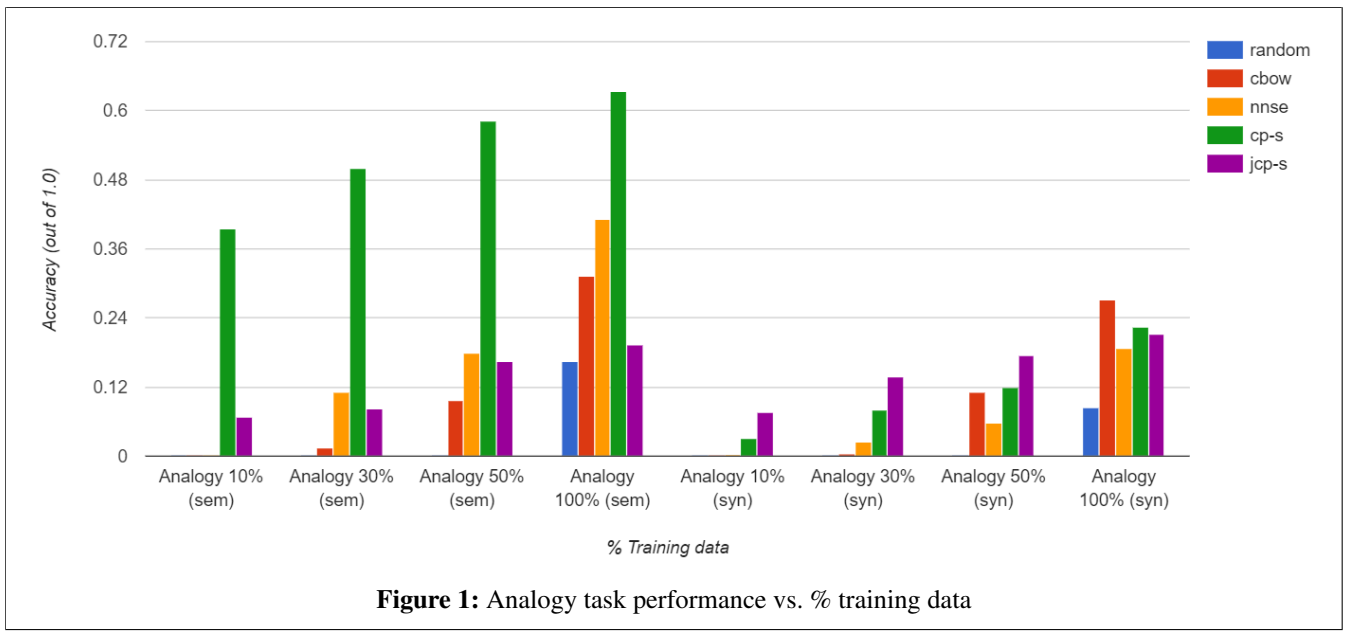
\createbigfig{analogytasks}{Analogy task performance vs. \% training data}{1.0}

\createbigfig{sentimenttasks}{Sentiment analysis task performance vs. \% training data}{1.0}

\renewcommand{\arraystretch}{1.0}
\begin{table}[!ht]    
    \centering
    \caption{Outlier Detection scores across all embeddings}
    \begin{tabular}{| c | c | c | c | c |} 
     \hline
     (Method) & \textbf{OD2 OPP} & \textbf{OD2 acc} & \textbf{OD3 OPP} & \textbf{OD3 acc}  \\ 
     \hline
    \textbf{Random} & 0.6123 & 0.2765 & 0.5345 & 0.1950 \\ 
    \textbf{CBOW} & 0.6542 & 0.3731 & 0.6162 & 0.3034 \\ 
    \textbf{NNSE} & 0.6998 & 0.4288 & 0.6292 & 0.3190 \\ 
    \textbf{CP-S} & \textbf{0.7078} & \textbf{0.4370} & \textbf{0.6741} & \textbf{0.3597} \\ 
    \textbf{JCP-S} & 0.7017 & 0.4242 & 0.6666 & 0.3201 \\ \hline
    \end{tabular}
    \label{tab:outlierscores}
\end{table} 

\textbf{Outlier Detection results.} The results are shown in \textbf{Table \ref{tab:outlierscores}}. The first thing to note is that CP-S outperforms the other methods across each Outlier Detection metric. Since the WikiSem500 dataset is semantically focused, performance at this task demonstrates the quality of semantic information encoded in our embeddings.

On OD2, the baselines perform more competitively with our CP Decomposition based models, but when OD3 is considered  our methods clearly excel. Since the tensor-based methods are trained directly on third order information and perform much better at OD3, we see that OD3 scores reflect the amount of third order information in a word embedding.
This is a validation of OD3, as our $3^{rd}$ order embeddings would naturally out perform $2^{nd}$ order embeddings at a task that requires third order information. Still, the superiority of our tensor-based embeddings at OD$2$ demonstrates the quality of the semantic information they encode.

\textbf{Supervised analogy results.} The results are shown in \textbf{Figure \ref{fig:analogytasks}}.
At the supervised semantic analogy task, CP-S vastly outperforms the baselines at all levels of training data, further signifying the amount of semantic information encoded by this embedding technique.

Also, when only 10\% of the training data is presented, our tensor methods are the only ones that attain nonzero performance -- even in such a limited data setting, use of CP-S's vectors results in nearly 40\% accuracy.
This phenomenon is also observed in the syntactic analogy tasks: our embeddings consistently outperform the others until 100\% of the training data is presented. 
These two observations demonstrate the accessibility of the information encoded in our word embeddings.
We can thus conclude that this relational information encoded in the tensor-based embeddings is more easily accessible than that of CBOW and NNSE.
Thus, our methods would likely be better suited for transfer learning to actual NLP tasks, particularly those in data-sparse settings.

\textbf{Sentiment analysis results.} The results are shown in \textbf{ Figure \ref{fig:sentimenttasks}}.
In this task, JCP-S is the dominant method across all levels of training data, but the difference is more obvious when training data is limited.
This again indicates that for this specific task the information encoded by our tensor-based methods is more readily available as that of the baselines. 
It is thus evident that exploiting both second and third order co-occurrence data leads to higher quality semantic information being encoded in the embedding.
At this point it is not clear why JCP-S so vastly outperforms CP-S at this task, but its superiority to the other strong baselines demonstrates the quality of information encoded by JCP-S.
This discrepancy is also illustrative of the fact that there is no single ``best word embedding'' \cite{simplesupervised} -- different embeddings encode different types of information, and thus should be used where they shine rather than for every NLP task.

\textbf{Word Similarity results.} 
\renewcommand{\arraystretch}{1.0}
\begin{table}[!ht]    
    \centering
    \caption{Word Similarity Scores (Spearman's $\rho$)}
    \begin{tabular}{| c | c | c |} 
     \hline
     (Method) & \textbf{MEN} & \textbf{MTurk} \\ 
     \hline
    \textbf{Random} & -0.028 & -0.150 \\ 
    \textbf{CBOW} & 0.601 & 0.498 \\ 
    \textbf{NNSE} & 0.717 & 0.686 \\ 
    \textbf{CP-S} & 0.630 & 0.631 \\ 
    \textbf{JCP-S} & 0.621 & 0.669 \\ \hline
    \end{tabular}
    \label{tab:wordsimscores}
\end{table} 

We show the results in \textbf{Table \ref{tab:wordsimscores}}.
As we can see, our embeddings very clearly outperform the random embedding at this task. They even outperform CBOW on both of these datasets. It is worth including these results as the word similarity task is a very common way of evaluating embedding quality in the literature. However, due to the many intrinsic problems with evaluating word embeddings using word similarity \cite{wordsim_problems}, we do not discuss this further. 


\section{Multiplicative Compositionality} \label{compositionality} 

We find that even though they are not explicitly trained to do so, our tensor-based embeddings capture polysemy information naturally through multiplicative compositionality.
We demonstrate this property qualitatively and provide proper motivation for it, leaving automated utilization to future work.

In our tensor-based embeddings, we found that one can create a vector that represents a word $w$ in the context of another word $w'$ by taking  the elementwise product ${\mathbf{v}}_w * {\mathbf{v}}_{w'}$. We call ${\mathbf{v}}_w * {\mathbf{v}}_{w'}$ a ``meaning vector'' for the polysemous word $w$.

For example, consider the word \textit{star}, which can denote a lead performer or a celestial body. We can create a vector for \textit{star} in the ``lead performer'' sense by taking the elementwise product $\mathbf{v}_{star} * \mathbf{v}_{actor}$. This produces a vector that lies near vectors for words related to lead performers and far from those related to \textit{star}'s other senses.

\renewcommand{\arraystretch}{1.15}
\begin{table*}[h!]
    \fontsize{7}{7}
    \centering
    \caption{Nearest neighbors (in cosine similarity) to elementwise products of word vectors}
    \makebox[\textwidth][c]{\begin{tabular}{| c | c | c | c |} 
     \hline
     \textbf{Composition} & \textbf{Nearest neighbors (CP-S)} & \textbf{Nearest neighbors (JCP-S)} & \textbf{Nearest neighbors (CBOW)}  \\ 
     \hline
     \textit{star} $*$ \textit{actor} & 
     
     \textit{oscar},
     \textit{award-winning},
     \textit{supporting}
      &
     
     \textit{roles},
     \textit{drama},
     \textit{musical}
      &
     
     \textit{DNA},
     \textit{younger},
     \textit{tip}
     
     \\
     \hline
     \textit{star} $+$ \textit{actor} & 
     
     \textit{stars},
     \textit{movie},
     \textit{actress}
      &
     
     \textit{actress},
     \textit{trek},
     \textit{picture}
      &
     
     \textit{actress},
     \textit{comedian},
     \textit{starred}
     
     \\ 
     \hline
     \textit{star} $*$ \textit{planet} & 
     
     \textit{planets},
     \textit{constellation},
     \textit{trek}
      &
     
     \textit{galaxy},
     \textit{earth},
     \textit{minor}
      &
     
     \textit{fingers},
     \textit{layer},
     \textit{arm}

     \\
     \hline
     \textit{star} $+$ \textit{planet} & 
     
     \textit{sun},
     \textit{earth},
     \textit{galaxy}
      &
     
     \textit{galaxy},
     \textit{dwarf},
     \textit{constellation}
      &
     
     \textit{galaxy},
     \textit{planets},
     \textit{earth}
     
     \\ 
     \hline
     \textit{tank} $*$ \textit{fuel} & 
     
     \textit{liquid},
     \textit{injection},
     \textit{tanks}
      &
     
     \textit{vehicles},
     \textit{motors},
     \textit{vehicle}
      &
     
     \textit{armored},
     \textit{tanks},
     \textit{armoured}

     \\
     \hline
     \textit{tank} $+$ \textit{fuel} & 
     
     \textit{tanks},
     \textit{engines},
     \textit{injection}
      &
     
     \textit{vehicles},
     \textit{tanks},
     \textit{powered}
      &
     
     \textit{tanks},
     \textit{engine},
     \textit{diesel}

     \\ 
     \hline
     \textit{tank} $*$ \textit{weapon} & 
     
     \textit{gun},
     \textit{ammunition},
     \textit{tanks}
      &
     
     \textit{brigade},
     \textit{cavalry},
     \textit{battalion}
      &
     
     \textit{persian},
     \textit{age},
     \textit{rapid}
     
          \\
     \hline
     \textit{tank} $+$ \textit{weapon} & 
     
     \textit{tanks},
     \textit{armor},
     \textit{rifle}
      &
     
     \textit{tanks},
     \textit{battery},
     \textit{batteries}
      &
     
     \textit{tanks},
     \textit{cannon},
     \textit{armored}

     
     
     
     
     
     
     
     
     
     
     
     
     
     
     
     
     \\ 
     \hline
    \end{tabular}}
    \label{tab:phrasevects}
\end{table*}

To motivate why this works, recall that the values in a third order PPMI tensor $\tensor{M}$ are given by:
\[
m_{ijk} = PPMI(w_i, w_j, w_k) \approx  \sum_{r=1}^R v_{ir} v_{jr} v_{kr} = \langle \mathbf{v}_{i} * \mathbf{v}_{j}, \mathbf{v}_{k} \rangle,
\]
where $\mathbf{v}_i$ is the word vector for $w_i$.
If words $w_i, w_j, w_k$ have a high PPMI, then $\langle \mathbf{v}_{i} * \mathbf{v}_{j}, \mathbf{v}_{k} \rangle$ will also be high, meaning $\mathbf{v}_{i} * \mathbf{v}_{j}$ will be close to $\mathbf{v}_{k}$ in the vector space by cosine similarity.

For example, even though \textit{galaxy} is likely to appear in the context of the word \textit{star} in in the ``celestial body'' sense, $\langle \mathbf{v}_{star} * \mathbf{v}_{actor}, \mathbf{v}_{galaxy} \rangle \approx$ PPMI(\textit{star, actor, galaxy}) is low whereas $\langle \mathbf{v}_{star} * \mathbf{v}_{actor}, \mathbf{v}_{drama} \rangle \approx$ PPMI(\textit{star, actor, drama}) is high. 
Thus , $\mathbf{v}_{star} * \mathbf{v}_{actor}$ represents the meaning of \textit{star} in the ``lead performer'' sense.

In \textbf{Table \ref{tab:phrasevects}} we present the nearest neighbors of multiplicative  and additive composed vectors for a variety of polysemous words. As we can see, the words corresponding to the nearest neighbors of the composed vectors for our tensor methods are semantically related to the intended sense \textit{both} for multiplicative and additive composition. In contrast, for CBOW, only additive composition yields vectors whose nearest neighbors are semantically related to the intended sense. Thus, our embeddings can produce complementary sets of polysemous word representations that are qualitatively valid whereas CBOW (seemingly) only guarantees meaningful additive compositionality. We leave automated usage of this property to future work.
    
    \section{Conclusion} \label{conclusion}
  Our key contributions are as follows: 
\begin{enumerate}[1.]
    \item 
    \textbf{Two novel tensor factorization based word embeddings.}
    We presented CP-S and JCP-S, which are word embedding techniques based on symmetric CP decomposition.
    We experimentally demonstrated that these embeddings outperform existing matrix-based techniques on a number of downstream semantic tasks when trained on the same data.
    
    \item 
    \textbf{A novel joint symmetric tensor factorization problem.}
    We introduced and utilized Joint Symmetric Rank-$R$ CP Decomposition to train JCP-S. 
    In this problem, multiple supersymmetric tensors must be decomposed using a \textit{single} rank-$R$ factor matrix. 
    This technique allows for utilization of both second and third order co-occurrence information in word embedding training.
    
    \item 
    \textbf{A new embedding evaluation metric to measure amount of third order information.}
    We produce a $3$-way analogue of Outlier Detection \cite{outlier1} that we call OD$3$.
    This metric evaluates the degree to which third order information is captured by a given word embedding.
    We demonstrated this by showing our tensor based techniques, which naturally encode third information, perform better at OD3 compared to existing second order models.
    
    \item 
    \textbf{Word vector multiplicative compositionality for polysemous word representation.} 
    We showed that our word vectors can be meaningfully composed \textit{multiplicatively} to create a ``meaning vector'' for each different sense of a polysemous word.
    This property is a consequence of the higher order information used to train our embeddings, and was empirically shown to be unique to our tensor-based embeddings.
\end{enumerate}
    
Tensor factorization appears to be a highly applicable and effective tool for learning word embeddings, with many areas of potential future work.
Leveraging higher order data in training word embeddings is useful for encoding new types of information and semantic relationships compared to models that are trained using only pairwise data.
This indicates that such techniques will prove useful for training word embeddings to be used in downstream NLP tasks.


\bibliography{sample}
  \bibliographystyle{aaai}

\end{document}